# Recognizing Facial Expressions in the Wild using Multi-Architectural Representations based Ensemble Learning with Distillation

Rauf Momin[1], Ali Shan Momin[1], Khalid Rasheed[1] and Muhammad Saqib[2]
[1] Department of Computing and Engineering Sciences, SZABIST, 75600 Karachi, Pakistan
[2] Faculty of Mechanical Science and Engineering, Institute of Materials Science, Technische Universität Dresden (TU Dresden), 01062 Dresden, Germany

Corresponding author: Rauf Momin (e-mail: raufmomin26@gmail.com), Muhammad Saqib (email: muhammad.saqib@tu-dresden.de)

**ABSTRACT** Facial expressions are the most common universal forms of body language. In the past few years, automatic facial expression recognition (FER) has been an active field of research. However, it is still a challenging task due to different uncertainties and complications. Nevertheless, efficiency and performance are yet essential aspects for building robust systems. We proposed two models, EmoXNet which is an ensemble learning technique for learning convoluted facial representations, and EmoXNetLite which is a distillation technique that is useful for transferring the knowledge from our ensemble model to an efficient deep neural network using label-smoothen soft labels for able to effectively detect expressions in real-time. Both of the techniques performed quite well, where the ensemble model (EmoXNet) helped to achieve 85.07% test accuracy on FER2013 with FER+ annotations and 86.25% test accuracy on RAF-DB. Moreover, the distilled model (EmoXNetLite) showed 82.07% test accuracy on FER2013 with FER+ annotations and 81.78% test accuracy on RAF-DB. Results show that our models seem to generalize well on new data and are learned to focus on relevant facial representations for expressions recognition.

**INDEX TERMS** Facial Expression Recognition, Affective Computing, Ensemble Learning, Deep Learning

## I. INTRODUCTION

Facial expressions are part of non-verbal ways of passing information from one individual to the other. They are extremely important to the social interaction of individuals. Hence, the ability to recognize facial expressions is really essential for better communication. Emotion recognition has emerged as an important research area, many works have been done in past for multiple modalities such as analyzing emotions from video, audio, text, and other physiological signals data. Communication is a real-time process so the level of uncertainty is much considerable, as the human-level performance for analyzing facial expressions is 65±5% accurate [1]. Hence, building an automated system to accurately detect and identify an expression in a real-time environment enables enormous areas of implementation. For example, developing systems for blind or visually impaired people, analyzing customer's food reviews, driver monitoring systems and, and other advanced human-computer interaction use cases.

Before the rise of deep neural networks and transfer learning, the traditional machine learning techniques with hand-crafted features were used such as dimensionality reduction techniques (e.g., PCA [2], LDA [3], SIFT descriptors [4], facial key-points, and landmarks [5]) followed by classifiers such as support vector machines (SVM), decision trees, or random forests. Nowadays with improvements in deep learning, many pre-trained CNN architectures have been proposed which is quite helpful for overcoming the need for manually extract features and able learn from fewer data with the help of transfer learning. Furthermore, developments with ensemble learning methods, are also very commonly used which allows to combine of several models and produces better predictions. There are many techniques available for ensemble learning such as stacking, averaging, voting, and many more. They are usually helpful for boosting up accuracy and obtaining better predictive performance.

To learn the key facial characteristics for the classification of seven universal expressions, we propose an ensemble learning model (EmoXNet) and an efficient distilled model (EmoXNetLite) for recognizing expressions with good performance and efficiency.

## II. RELATED WORKS

Facial Expression Recognition has been widely studied and an active field of research. In past, as a part of the FER Challenge which was conducted using the FER2013 dataset [1]. The top three teams all used convolutional neural networks. The winner of the challenge, Yichuan Tang [6] used convolutional neural networks with linear support vector machines. Results from this competition suggest that convolutional neural networks are capable to outperform hand-crafted features. Another paper, by Pramerdorfera and Kampel [7] describes their approach by forming an ensemble of 8 CNN's and achieving 75.2% test accuracy. Riaz et al. [8] presented their custom CNN architecture named eXnet based on a parallel feature extraction approach for emotion recognition in the wild. Moreover, there have been several developments in other techniques also for facial expression recognition such as a spatiotemporal bilinear network [9] using facial landmarks, graph convolutional networks [10], etc. Some of them have their own limitations with different aspects. The objective of this paper is to focus and deal with the key limitations from previous researches, like real-time competence and overall system performance which are usually the significant aspects for building efficient robust systems.

## III. METHODOLGY

For learning discriminative and robust facial representations and also to achieve better overall performance. We developed an ensemble learning technique for learning convoluted facial representations using multi-layer perceptron as a meta-classifier. As shown below in FIGURE 1.

**FIGURE 1** Proposed (EmoXNet) model architecture diagram with different feature extractors blocks.

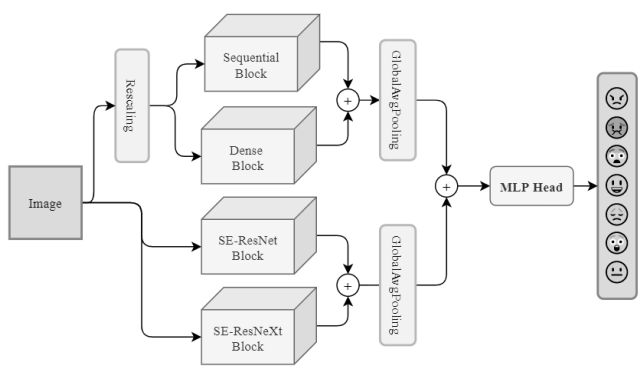

In this approach, we ensemble the features extracted from various backbones such as sequential block, dense block, se-resnet block, se-resnext block which refers to vgg16 [11], densenet121 [12], squeeze and excitation resnet34 [13], [14], and squeeze and excitation resnext50 [14] respectively, and using the multi-layer perceptron as our meta-classifier head for making the final predictions from concatenated average pooled features with softmax as the last layer.

Moreover, for dealing with requirements such as efficient and fast recognition of expressions in real-time and resource constraints systems. We proposed a distillation technique using categorical cross-entropy loss function with label smoothing for transferring knowledge learned from our ensemble model to the lightweight neural network. The intuition behind our distillation approach is based on the idea provided in the knowledge distillation [15] paper which used temperature as a variable for their custom loss function to provide more information in predicted labels for the student model. To understand the problem well, we initially trained our model using one-hot encoded labels with categorical cross-entropy as a loss function and usual softmax activation output at the last layer. Which can be mathematically denoted as follows,

$$\mathcal{L} = -\sum_{i=1}^{N} y_i \cdot \log(f(\hat{y})_i) \qquad (1)$$

$$f(\hat{y})_i = \frac{e^{\hat{y}_i}}{\sum_{j=1}^{N} e^{\hat{y}_j}} \qquad (2)$$

The Eq. (1) computes categorical cross-entropy loss using ground-truth labels ($y_i$) and the log of softmax activation function applied to trained logits ($\hat{y}_i$) with summation over a total number of classes ($N$). The Eq. (2) shows the mathematical representation of the softmax activation function where logits are passed as input. Subsequently, initial error analysis showed, mislabeled images seemed much more common problem for this task. For tackling this issue, we decided to incorporate the label smoothing with categorical cross-entropy loss. Which is generally used as the regularization method to deal with overfitting and help the model generalize well. The Eq. (3) shown below, calculates smoothen labels from softmax output. Where $\alpha$ is denoted as the label smoothing factor which is a hyperparameter that regulates the amount of smoothing. We used the smoothing factor of 0.1 to relax confidence on the labels and able to hold more information which will be useful for our lightweight model. The label-smoothen loss function can be formulated as,

$$\hat{y}_{ls} = (1 - \alpha) * f(\hat{y})_i + \frac{\alpha}{N} \qquad (3)$$

At the time of inference, we also tested out test time augmentations technique. As it's common practice to use data augmentations for improving robustness and preventing overfitting while training a model. Test-Time Augmentation (TTA) is a technique that can help to boost a model's prediction accuracy during the inference phase by applying augmentations for a certain number of steps and then averaging the output predictions achieved from each step. We used TTA with 10 steps which showed improvement in the accuracy of the model.

## IV. EXPERIMENTS

For evaluating the effectiveness of the model. There are various datasets are accessible, we used two different datasets to conduct our experiments and in order to comparatively analyze different CNN architectures.

FER-2013 [1] is publicly available dataset and widely used in many research papers with around 35,887 grayscale 48x48 pixels images. For better performance and accuracy, we used updated and improved FER+ annotations [16]. The updated dataset was around 35,249 images considering following seven classes (angry, disgust, fear, happy, sad, surprise and neutral).

RAF-DB [17] is another large-scale facial expression database with around with around 30,000 facial images. We used single-labeled images from RAF-DB which was around 15,339 RGB 100x100 pixels images. For consistency we converted them to 48x48 resized images.

FIGURE 2 Visualizing data augmentations on sample image.

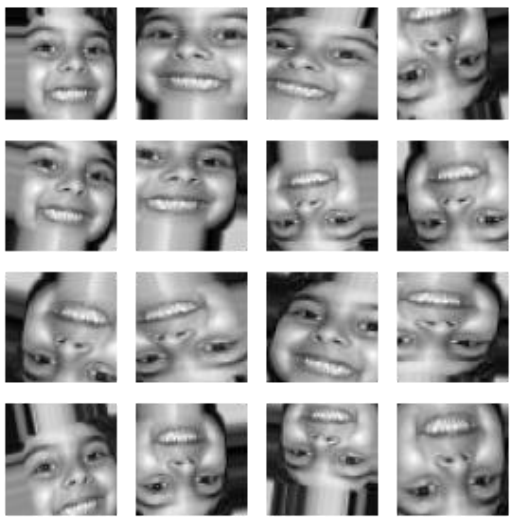

FIGURE 2 shows the visualization of list of data augmentations techniques (horizontal flip, vertical flip, rotation range, width shift range, height shift range, shear range, zoom range, and fill mode to nearest) applied to single image. Furthermore, we performed training on architectural different pre-trained [18] models such as sequential, residual, squeeze and excitation, densely connected and many others which helped us to comparatively examine the facial features.

The experiments were conducted using Nesterov SGD with momentum as an optimizer, where momentum was set to 0.9, initial learning rate set to 0.01, and decay of 0.0001. Correspondingly, the models are trained on both the datasets for 100 epochs and a batch size of 128. We also tried experimenting with other optimizers like Adam and AMSGrad for finding an optimal solution but after comparing initial results from them it was clear that SGD was able to generalize and converge well and was also able to obtain better results than other optimizers.

## V. RESULTS AND ANALYSIS

The results from our experiments and analysis on different models are shown in TABLE 1 and TABLE 2 where accuracy was set as a metric for evaluation and also used as a monitoring parameter by early stopping and learning rate scheduler callback for preventing overfitting and for finding global minima respectively.

TABLE 1 Results on FER2013 dataset with FER+ annotations.

| Model | Accuracy |
|---|---|
| vgg16 | 79.68 % |
| densenet121 | 82.32 % |
| efficientnet-B0 | 78.78 % |
| resnet34 | 81.16 % |
| se-resnet34 | 80.31 % |
| resnext50 | 81.84 % |
| se-resnext50 | 82.77 % |
| *EmoXNet (Ours)* | *85.07 %* |
| *EmoXNetLite (Ours)* | *86.25 %* |

TABLE 2 Results on Real-world Affective Faces Database.

| Model | Accuracy |
|---|---|
| vgg16 | 77.38 % |
| densenet121 | 78.75 % |
| efficientnet-B0 | 70.21 % |
| resnet34 | 76.56 % |
| se-resnet34 | 74.58 % |
| resnext50 | 77.41 % |
| se-resnext50 | 76.99 % |
| *EmoXNet (Ours)* | *82.07 %* |
| *EmoXNetLite (Ours)* | *81.78 %* |

TABLE 1 and TABLE 2, show the comparison among various deep convolutional feature extractors with the same classification head with our two proposed models (EmoXNet which is the ensemble model, and EmoXNetLite which is distilled model). As results clearly suggest that, the proposed models are best performing and stable among all methods. The perception behind the model's achieving robust performance to unseen data is based on the diverse representations, which are learned from different architectures which helped to highlight various key characteristics. To further analyze the performance, the class-wise F1 scores of these two models are shown below,

TABLE 3 F1-Scores on FER2013 with FER+ annotations.

| Class | EmoXNet | EmoXNetLite |
|---|---|---|
| Angry | 0.79 | 0.79 |
| Disgust | 0.58 | 0.67 |
| Fear | 0.60 | 0.59 |
| Happy | 0.93 | 0.92 |
| Sad | 0.68 | 0.68 |
| Surprise | 0.88 | 0.88 |
| Neutral | 0.87 | 0.89 |

TABLE 4 F1-Scores on Real-world Affective Faces Database.

| Class | EmoXNet | EmoXNetLite |
|---|---|---|
| Angry | 0.72 | 0.70 |
| Disgust | 0.47 | 0.50 |
| Fear | 0.58 | 0.70 |
| Happy | 0.92 | 0.90 |
| Sad | 0.78 | 0.75 |
| Surprise | 0.80 | 0.81 |
| Neutral | 0.79 | 0.81 |

After looking at the results from TABLE 3 and TABLE 4, it can be clearly interpreted that classes disgust and fear are hard to predict because the support count for both classes is relatively low than others. While the most common expressions like happy, surprise, and neutral are easy to predict, reasons include visually distinguishing features and better support count which enables the model to predict efficiently.

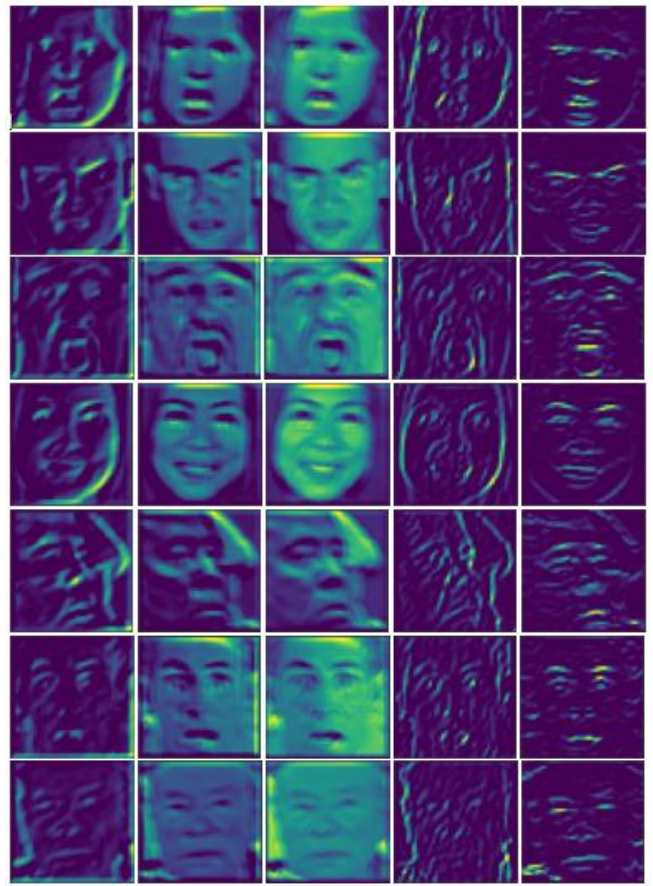

FIGURE 3 Class Activation Visualization of classes (Angry, Disgust, Fear, Happy, Sad, Surprise, Neutral) left to right.

As seen in FIGURE 3, from the class activation visualizations the dominating representations learn from the model are of the edges and key facial areas which is certainly very useful for identifying the expressions in a frontal-face image. From visualizations, we can also notice the feature representation for classes with insufficient support count is a bit distorted to highlight the features whereas, classes with a sufficient amount of data have better representations of the activation maps. Problems such as class imbalance and mislabeled images are the major issues for this task.

## VI. CONCLUSION AND FUTURE WORK

In this paper, we reviewed the different CNN architectures for recognizing expressions. We also developed an ensemble approach that helped to encode learned representations in MLP and also helpful to obtain the state-of-the-art testing performance on two different benchmark datasets. Additionally, we demonstrated the distillation technique for passing knowledge from a large ensemble model to lightweight efficient neural networks for enabling emotion recognition with low latency for real-time systems. Furthermore, with the help of class activation visualizations, we showed that models are learned to focus on relevant key facial features for detecting the expressions. In the future, we planned to work on the model compression and optimization techniques such as quantization, to optimize the set of parameters for able to detect more effectively for low-end and edge devices. Simultaneously, another objective will be to deal with the quality of the data labeling by employing semi-supervised learning techniques which will indeed help to improve overall performance.

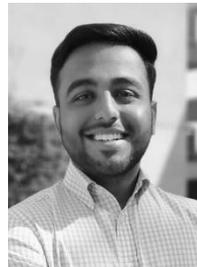

**RAUF MOMIN** was born in Karachi, Pakistan, in 1999. He received the B.S. degree in Computer Science from SZABIST, Karachi, Pakistan. He has adequate research and academic experience. In his free time, he enjoys exploring new stuff related to his field and participating in Kaggle competitions. He's broadly interested in deep learning and pattern recognition. Specifically, his research interests include deep generative models, image processing, and applications of AI for healthcare and medicine.

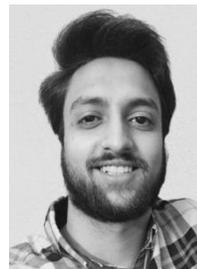

**ALI SHAN MOMIN** graduated from Shaheed Zulfiqar Ali Bhutto Institute of Science and Technology, Karachi with a bachelor's degree in computer science. In this domain, he has sufficient academic experience. He is very enthusiastic and motivated to learn new technologies. He is very passionate about new trends in market and has very good academic background. Web and mobile app development, machine learning and data science, are among his areas of interest.

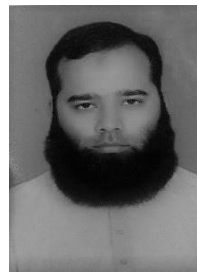

**KHALID RASHEED** received the Bachelor's Degree from Karachi University, Pakistan, and the MS. degree in computer science from SZABIST, Karachi, Pakistan. He is currently pursuing a Ph.D. degree in computer science at SZABIST, Karachi, Pakistan. He has gained academic and industrial experience in his professional career. His research interest includes the machine learning, computer vision, image processing, data sciences and cyber security.

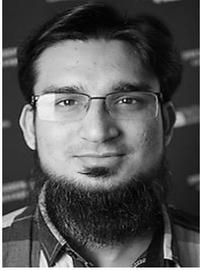

**MUHAMMAD SAQIB** received the B.S. degree in biomedical engineering from the Sir Syed University of Engineering and Technology, Pakistan, and the M.S. degree in electronic engineering from Hamdard University, Karachi, Pakistan. He is currently pursuing the Ph.D. degree in biomedical engineering at Fraunhofer IKTS, Dresden, Germany and TU Dresden, Dresden, Germany. He has gained academic and industrial experience in his professional career. His research interest includes the stent degradation, bioinstrumentation, biomaterials, biomechanics, corrosion. He has two patents and several peer-reviewed conference and journal publications. Mr. Saqib got the HEC Pakistan/DAAD scholarship for his PhD studies.